\documentclass[11pt,a4paper]{article}
\usepackage{authblk}
\usepackage[hyperref]{emnlp-ijcnlp-2019}
\usepackage{times}
\usepackage{latexsym}
\usepackage{url}

\usepackage{amsmath}
\usepackage{amssymb}
\usepackage{graphicx}
\usepackage{paralist}
\usepackage{multirow}
\usepackage{nicefrac}
\usepackage{rotating}
\usepackage{float}
\usepackage{ltablex}

\DeclareMathOperator*{\argmin}{arg\,min}
\newcommand\sbullet[1][.5]{\mathbin{\vcenter{\hbox{\scalebox{#1}{$\bullet$}}}}}

\aclfinalcopy

\title{Energy-based Self-attentive Learning of Abstractive Communities for Spoken Language Understanding}

\author[1,2]{Guokan Shang}
\author[1]{Antoine J.-P. Tixier}
\author[1,3]{\\Michalis Vazirgiannis}
\author[2]{Jean-Pierre Lorr\'e}
\affil[1]{\'Ecole Polytechnique, $^\mathrm{2}$Linagora, $^\mathrm{3}$AUEB}

\date{}

\begin{document}
\maketitle
\begin{abstract}
Abstractive community detection is an important spoken language understanding task, whose goal is to group utterances in a conversation according to whether they can be jointly summarized by a common abstractive sentence. 
This paper provides a novel approach to this task. We first introduce a neural contextual utterance encoder featuring three types of self-attention mechanisms. We then train it using the siamese and triplet energy-based meta-architectures.
Experiments on the AMI corpus show that our system outperforms multiple energy-based and non-energy based baselines from the state-of-the-art. 
Code and data are publicly available\footnote{\tiny\url{https://bitbucket.org/guokan_shang/abscomm}}.
\end{abstract}

\section{Introduction} \label{sec:introduction}
Today, large amounts of digital text are generated by spoken or written conversations, let them be human-human (customer service, multi-party meetings) or human-machine (chatbots, virtual assistants). Such text comes in the form of transcriptions. A transcription is a list of time-ordered text fragments called \textit{utterances}. Unlike sentences in traditional documents, utterances are frequently associated with meta-information in the form of discourse features such as speaker ID/role, dialogue act, etc. Utterances are also often ill-formed, incomplete, and ungrammatical, due to the nature of spontaneous communication.

\begin{figure}[ht]
\centering
\captionsetup{size=small}
\includegraphics[scale=0.33]{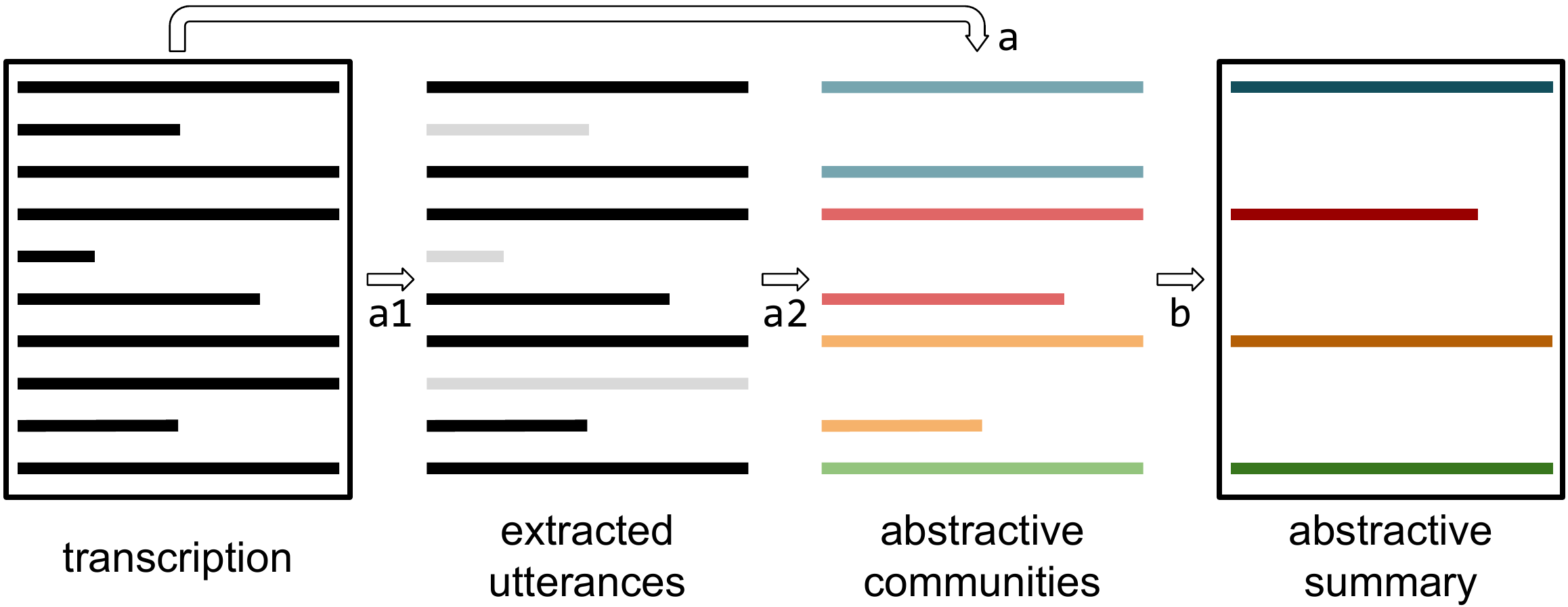}
\caption{Abstractive Community Detection (subtask $a$) is the first step towards summarizing a conversation.}
\label{fig:acd}
\end{figure}

\begin{figure*}[ht]
\centering
\captionsetup{size=small}
\includegraphics[scale=0.78]{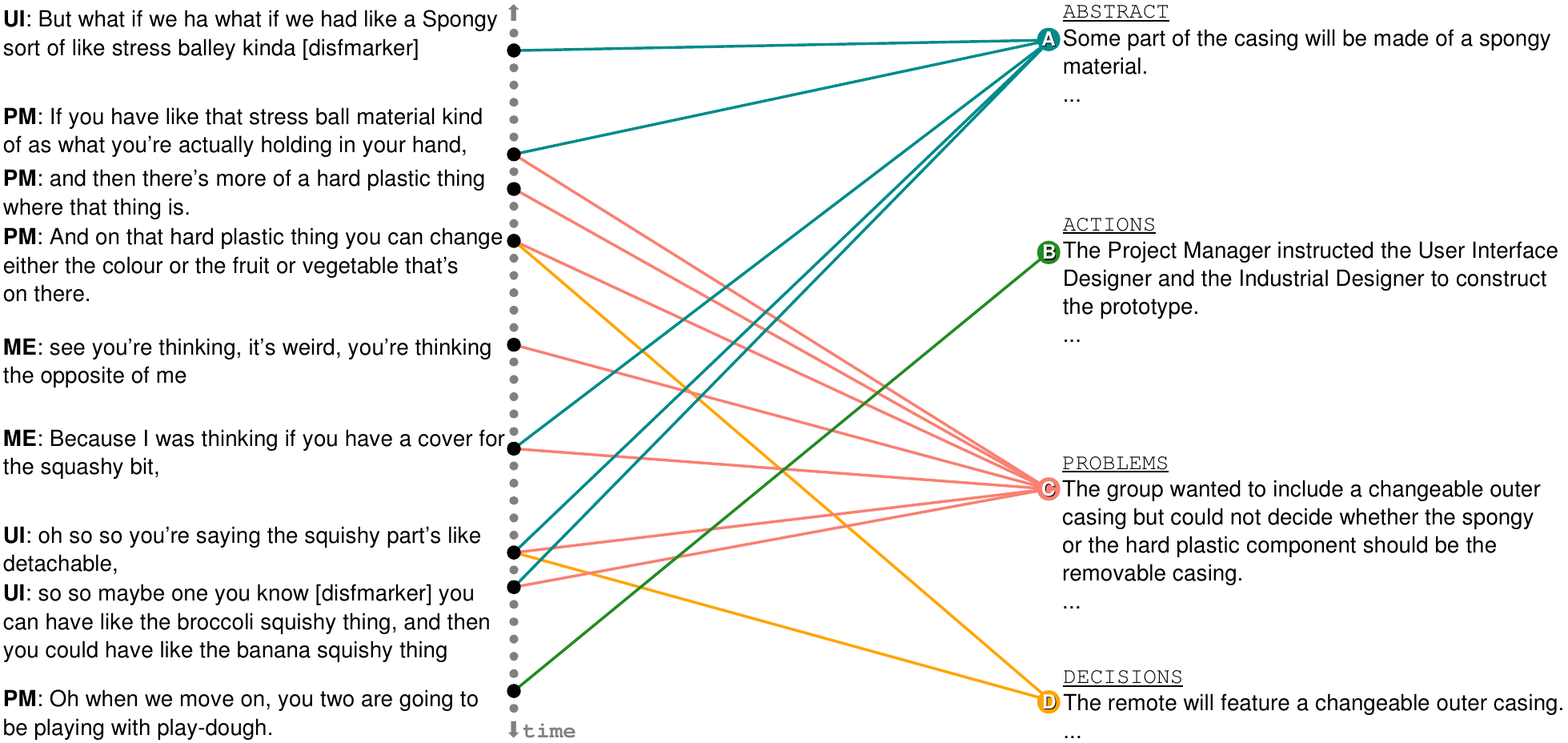}
\caption{Example of ground truth human annotations from the ES2011c AMI meeting. Successive grey nodes on the left denote utterances in the transcription, where black nodes correspond to the utterances judged important (summary-worthy). Sentences from the abstractive summary are shown on the right. All utterances linked to the same abstractive sentence form one community. Speaker roles are PM: project manager, ME: marketing expert, UI: user interface designer.}
\label{fig:bipartite}
\end{figure*}

\noindent Abstractive summarization of conversations is an open problem in NLP. It requires the machine to gain a high-level understanding of the dialogue, in order to extract useful information and turn it into meaningful abstractive sentences.
Previous work \citep{mehdad2013abstractive, oya2014template, banerjee2015generating, shang2018unsupervised} decomposes this task into two subtasks \texttt{a} and \texttt{b} as shown in Fig. \ref{fig:acd}. 

Subtask \texttt{a}, or \textit{Abstractive Community Detection} (ACD), is the focus of this paper. It consists in grouping utterances according to whether they can be jointly summarized by a common abstractive sentence \citep{murray2012using}. Such groups of utterances are called \textit{abstractive communities}. Once they are obtained, an abstractive sentence is generated for each group (subtask \texttt{b}), thus forming the final summary. ACD includes, but is a more general problem than, topic clustering. Indeed, as shown in Fig. \ref{fig:bipartite}, communities should capture more complex relationship than simple semantic similarity. Also, two utterances may be part of the same community even if they are not close to each other in the transcription. Finally, a given utterance may belong to more than one community, which results in overlapping groupings (e.g., A and D in Fig. \ref{fig:bipartite}), or be a community of its own, i.e., a singleton community (B in Fig. \ref{fig:bipartite}).

In this paper, we depart from previous work and argue that the ACD subtask should be broken down into two steps, \texttt{a1} and \texttt{a2} in Fig. \ref{fig:acd}. That is, summary-worthy utterances should first be extracted from the transcription (\texttt{a1}), and then, grouped into abstractive communities (\texttt{a2}). This $a1 \rightarrow a2 \rightarrow b$ process is more consistent with the way humans treat the summarization task. E.g., during the creation of the AMI corpus \citep{mccowan2005ami}, annotators were first asked to extract summary-worthy utterances from the transcription, and then to link the selected utterances to the sentences in the abstractive summary (links in Fig. \ref{fig:bipartite}), i.e., create communities. Abstractive summaries comprise four sections: \texttt{ABSTRACT}, \texttt{ACTIONS}, \texttt{PROBLEMS}, and  \texttt{DECISIONS}.

Step \texttt{a1} plays an important filtering role, since in practice, only a small part of the original utterances are used to construct the abstractive communities (17\% on average for AMI).
However, this step is closely related to \textit{extractive summarization}, which has been extensively studied in the conversational domain  \citep{murray2005extractive, garg2009clusterrank, tixier2017combining}.

Rather, we focus in this paper on the rarely explored \texttt{a2} \textit{utterance clustering} step, which we think is an important spoken language understanding problem, as it plays a crucial role of bridge between two major types of summaries: extractive and abstractive.

\section{Departure from previous work}\label{sec:related}
Prior work performed ACD either in a supervised \citep{murray2012using, mehdad2013abstractive} or unsupervised way \citep{oya2014template, banerjee2015generating, singla2017automatic, shang2018unsupervised}. 

In the supervised case, \citet{murray2012using} train a logistic regression classifier with handcrafted features to predict extractive-abstractive links, then build an utterance graph whose edges represent the binary predictions of the classifier, and finally apply an overlapping community detection algorithm to the graph.
\citet{mehdad2013abstractive} add to the previous approach by building an entailment graph for each community, where edges are entailment relations between utterances, predicted by a SVM classifier trained with handcrafted features on an external dataset. The entailment graph allows less informative utterances to be eliminated from each community. 

On the other hand, unsupervised approaches to ACD do not make use of extractive-abstractive links.
\citet{oya2014template,banerjee2015generating,singla2017automatic} assume that disjoint topic segments \citep{galley2003discourse, eisenstein2008bayesian} align with abstractive communities, while
\citet{shang2018unsupervised} use the classical vector space representation with TF-IDF weights, and apply $k$-means to the LSA-compressed utterance-term matrix.

To sum up, prior ACD methods either train multiple models on different labeled datasets and heavily rely on handcrafted features, or are incapable of capturing the complicated structure of abstractive communities described in the introduction.

Motivated by the recent success of energy-based approaches to similarity learning tasks such as face verification \citep{schroff2015facenet} and sentence matching \citep{mueller2016siamese}, we introduce in this paper a novel utterance encoder, and train it within the siamese \cite{chopra2005learning} and triplet \cite{hoffer2015deep} energy-based meta-architectures. Our final network is able to accurately capture the complexity of abstractive community structure, while at the same time, it is trainable in an end-to-end fashion without the need for human intervention and handcrafted features. Our contributions are threefold:

\noindent $\bullet$ we formalize ACD, a crucial subtask for abstractive summarization of conversations, and publicly release a version of the AMI corpus preprocessed for this subtask, to foster research on this topic,

\noindent $\bullet$ we propose one of the first applications of energy-based learning to spoken language understanding,

\noindent $\bullet$ we introduce a novel utterance encoder featuring three types of self-attention mechanisms and taking contextual and temporal information into account.

\section{Energy-based learning}\label{sec:energy}
Energy-Based Modeling (EBM) \cite{lecun2005loss,lecun2006tutorial} is a unified framework that can be applied to many machine learning problems. In EBM, an energy function assigns a scalar called \textit{energy} to each pair of random variables $(X,Y)$. The energy can be interpreted as the incompatibility between the values of $X$ and $Y$.
Training consists in finding the parameters $W^\ast$ of the energy function $E_W$ that, for all $(X^i,Y^i)$ in the training set $\mathcal{S}$ of size $P$, assign low energy to compatible (correct) combinations and high energy to all other incompatible (incorrect) ones. This is done by minimizing a \textit{loss functional}\footnote{the loss \textit{functional} is passed the output of the energy function, unlike a loss \textit{function} which is directly fed the output of the model.} $\mathcal{L}$:
\begin{equation}
    W^\ast=\argmin_{W\in\mathcal{W}}\mathcal{L}(E_W(X,Y),\mathcal{S})
\end{equation}
\noindent For a given $X$, prediction consists in finding the value of $Y$ that minimizes the energy.
\begin{figure}[t]
    \centering
    \captionsetup{size=small}
    \includegraphics[scale=0.36]{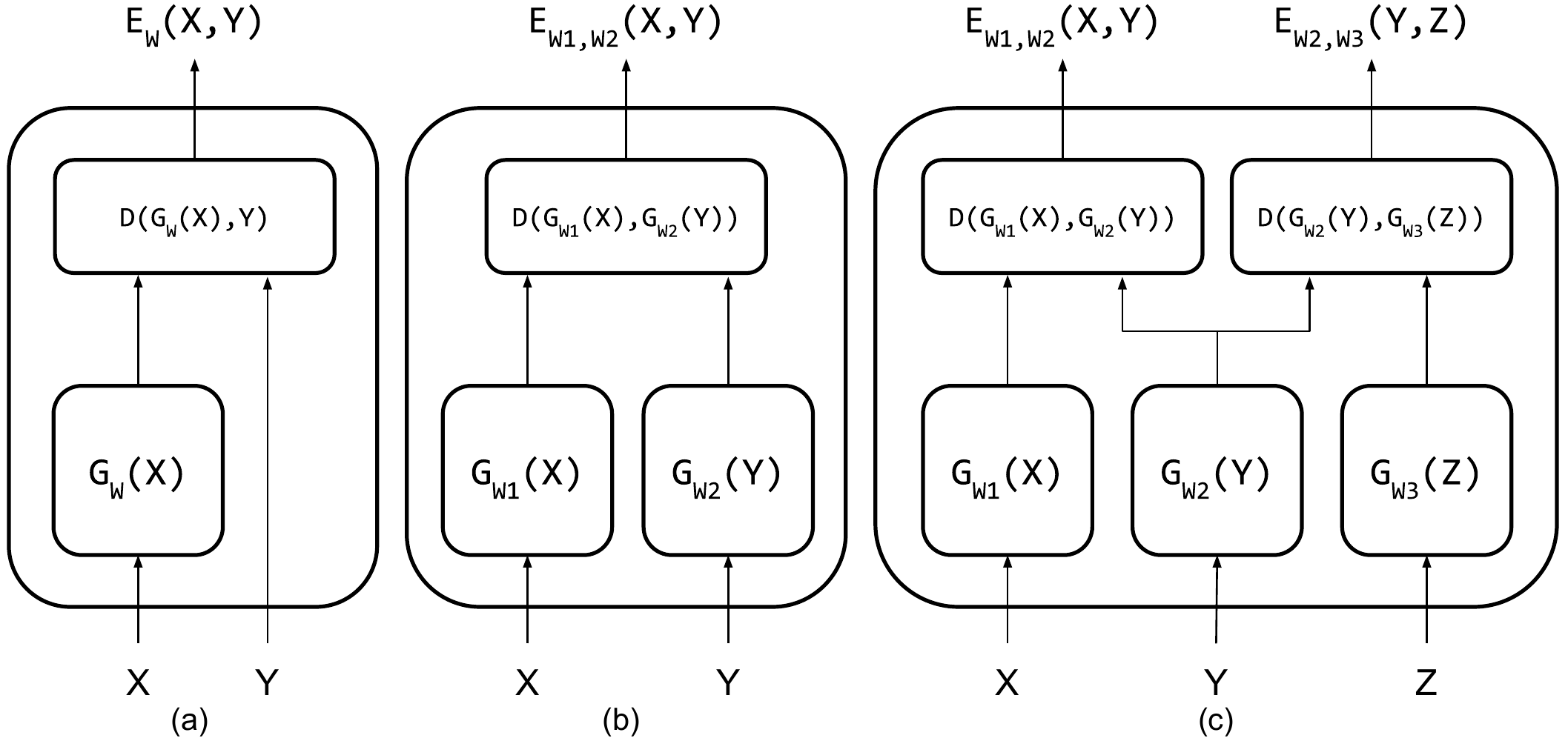}
	\caption{Three EBM architectures. When all $G$s and $W$s are equal, (b) and (c) correspond to the siamese/triplet cases.
	}
	\label{fig:ebms}
\end{figure}

\subsection{Single architecture}
In the EBM framework, a regression problem can be formulated as shown in Fig. \ref{fig:ebms}a, where the input $X$ is passed through a regressor model $G_W$ and the scalar output is compared to the desired output $Y$ with a dissimilarity measure $D$ such as the squared error. Here, the energy function is the loss functional to be minimized.

{\small
\setlength{\abovedisplayskip}{0pt}
\setlength{\belowdisplayskip}{0pt}
\begin{equation}
    \mathcal{L} = \frac{1}{P} \sum_{i=1}^P E_W(X^i,Y^i)= \frac{1}{P} \sum_{i=1}^P\|G_W(X^i)-Y^i\|^2
\end{equation}
}

\subsection{Siamese architecture}\label{sub:siam}
In the regression problem previously described, the dependence between $X$ and $Y$ is expressed by a direct mapping $Y=f(X)$, and there is a single best $Y^\ast$ for every $X$. However, when $X$ and $Y$ are not in a predictor/predictand relationship but are exchangeable instances of the same family of objects, there is no such mapping. E.g., in paraphrase identification, a sentence may be similar to many other ones, or, in language modeling, a given $n$-gram may be likely to be followed by many different words. 

Thereby, \citet{lecun2006tutorial} introduced EBM for \textit{implicit regression} or \textit{constraint satisfaction} (see Fig. \ref{fig:ebms}b), in which a constraint that $X$ and $Y$ must satisfy is defined, and the energy function measures the extent to which that constraint is violated:
\begin{equation}
    E_{W_1,W_2}(X,Y) = D(G_{W_1}(X), G_{W_2}(Y))
\end{equation}

\noindent where $G_{W_2}$ and $G_{W_1}$ are two functions parameterized by $W_1$ and $W_2$. When $G_{W_1}=G_{W_2}$ and $W_1=W_2$, we obtain the well-known siamese architecture \cite{bromley1994signature, chopra2005learning}, which has been applied with success to many tasks, including sentence similarity \cite{mueller2016siamese}. 

Here, the constraint is determined by a collection-level set of binary labels $\{C^i\}_{i=1}^{P}$. E.g., $C^i=0$ indicates that $(X^i, Y^i)$ is a \textit{genuine} pair (e.g., two paraphrases), while $C^i=1$ indicates that $(X^i, Y^i)$ is an \textit{impostor} pair (e.g., two sentences with different meanings).

The function $G_W$ projects objects into an embedding space such that the defined dissimilarity measure $D$ (e.g., Euclidean distance) in that space reflects the notion of dissimilarity in the input space. Thus, the energy function can be seen as a metric to be learned.

We experiment with various deep neural network encoders as $G_W$, and, following \cite{mueller2016siamese}, we adopt the exponential negative Manhattan distance as dissimilarity measure  and the mean squared error as loss functional:

{\small
\setlength{\abovedisplayskip}{0pt}
\setlength{\belowdisplayskip}{0pt}
\begin{align}
E_{W}(X,Y) &= 1-\exp(-\|G_{W}(X) - G_{W}(Y)\|_1) \\
\mathcal{L} &= \frac{1}{P} \sum_{i=1}^P\|E_{W}(X^i,Y^i)-C^i\|^2 \label{eq:siamese}
\end{align}
}

\subsection{Triplet architecture}\label{sub:trip}
The triplet architecture \cite{schroff2015facenet, hoffer2015deep, wang2014learning}, as can be seen in Fig. \ref{fig:ebms}c, is a direct extension of the siamese architecture that takes as input a triplet $(X,Y,Z)$ in lieu of a pair $(X,Y)$. $X$, $Y$, and $Z$ are referred to as the \textit{positive}, \textit{anchor}, and \textit{negative} objects, respectively. $X$ and $Y$ are similar, while both being dissimilar to $Z$. Learning consists in jointly minimizing the positive-anchor energy $E_{W}(X^i,Y^i)$ while maximizing the anchor-negative energy $E_{W}(Y^i,Z^i)$.

Here, we use the \textit{softmax triplet loss} \citep{hoffer2015deep} as our loss functional:
\begin{align}
\mathcal{L} = &\frac{1}{2P} \sum_{i=1}^P \big(\|ne^+-0\|^2 + \|ne^--1\|^2\big)\\
ne^+ &= \frac{e^{E_{W}(X^i,Y^i)}}{e^{E_{W}(X^i,Y^i)}+e^{E_{W}(Y^i,Z^i)}}\\
ne^- &= \frac{e^{E_{W}(Y^i,Z^i)}}{e^{E_{W}(X^i,Y^i)}+e^{E_{W}(Y^i,Z^i)}}
\end{align}

\noindent where $ne$ stands for normalized energy, and the dissimilarity measure is the Euclidean distance, i.e., {\small$E_{W}(X^i,Y^i) = \|G_{W}(X^i) - G_{W}(Y^i)\|_2$}. Essentially, the softmax triplet loss is the mean squared error between the normalized energy vector $[ne^+,ne^-]$ and $[0,1]$.

\subsection{On our choice of loss functionals}
\begin{figure}[ht]
\centering
\captionsetup{size=small}
\includegraphics[scale=1.3]{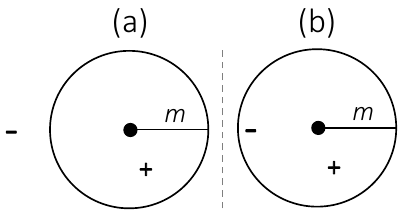}
\caption{$\sbullet[.75]$, -, + denote anchor, negative, and positive.}
\label{fig:margin}
\end{figure}

The softmax triplet loss (STL) performed better in our experiments than the margin-based loss used in \cite{schroff2015facenet} and \cite{wang2014learning}. One of the reasons may be that STL is able to capture a finer notion of distance. Indeed, with a margin-based loss, the Euclidean distance between the anchor and the negative (let us compactly denote it as $d^-$) need to satisfy $d^->d^+ + m$, where $m$ is the margin (see Fig. \ref{fig:margin}a). In other words, the distance between the positive and the negative is at least $m$ (when all three points are aligned).

\noindent However, the objective of STL is simply $d^->d^+$, without imposing an absolute lower bound on the distance between positives and negatives (i.e., only the distance ratio is of interest, see Fig. \ref{fig:margin}b), which gives more freedom to the model.

\noindent For consistency, we also adopt a margin-free loss functional for siamese (MSE, see Eq. \ref{eq:siamese}). It also performed better than the traditional contrastive loss \cite{chopra2005learning, neculoiu2016learning} in early experiments.

\subsection{Sampling procedures} \label{subsec:sampling}
We sample tuples from the ground truth abstractive communities to train our utterance encoder $G_W$ (see section \ref{sec:proposed}) under the siamese and triplet meta-architectures as follows.

\noindent \textbf{Pair sampling}.
All utterances belonging to the same community are paired as genuine pairs, while impostor pairs are any two utterances coming from different communities.

\noindent \textbf{Triplet sampling}.
Utterances from the same community provide positive and anchor items, while the negative item is taken from any other community.

\begin{figure*}[ht]
\centering
\captionsetup{size=small}
\includegraphics[scale=0.65]{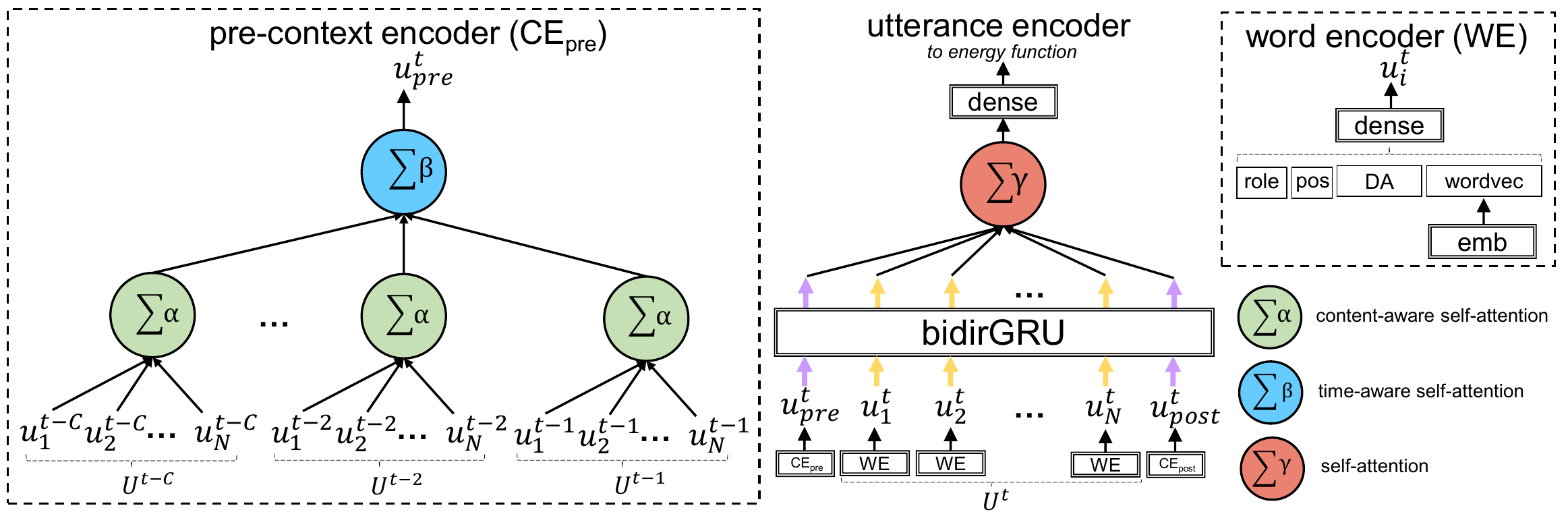}
\caption{Our proposed utterance encoder. Only the pre-context encoder is shown. $C$ is the context size.}
\label{fig:architecture}
\end{figure*}

\section{Proposed utterance encoder}\label{sec:proposed}
\noindent\textbf{Notation}.
The \textit{time} $t$ (as superscript) denotes the position of a given utterance in the conversation of length $T$, and the \textit{position} $i$ (as subscript) denotes the position of a token within a given utterance of length $N$. E.g., $\mathbf{u}_{1}^t$ is the representation of the first token of $\mathbf{U}^t$, the $t^{th}$ utterance in the transcription. Upper and lower case are used for matrices and vectors. Vectors are distinguished from floats by using boldface.

\subsection{Word encoder} \label{subsec:word_encoder}
As shown in the upper right corner of Fig. \ref{fig:architecture}, we obtain $\mathbf{u}_i^t$ by concatenating the pre-trained vector of the corresponding token with the discourse features of $\mathbf{U}^t$ (role, position and dialogue act), and passing the resulting vector to a dense layer.

\subsection{Utterance encoder}\label{sub:uttenc}
As shown in the center of Fig. \ref{fig:architecture}, we represent $\mathbf{U}^t$ as a sequence of $N$ $d$-dimensional token representations $\big\{\mathbf{u}_{1}^t,\dots ,\mathbf{u}_{N}^t\big\}$. 
In addition, because there is a strong time dependence between utterances (see Fig. \ref{fig:bipartite}), we inform the model about the preceding and following utterances when encoding $\mathbf{U}^t$. To accomplish this, we prepend (resp. append) to $\mathbf{U}^t$ a context vector containing information about the previous (resp. next) utterances, finally obtaining $\mathbf{U}^t=\big\{\mathbf{u}_{\mathrm{pre}}^t,\mathbf{u}_{1}^t,\dots ,\mathbf{u}_{N}^t,\mathbf{u}_{\mathrm{post}}^t\big\} \in \mathbb{R}^{(N+2) \times d}$. We then use a non-stacked bidirectional Recurrent Neural Network (RNN) with Gated Recurrent Units (GRU) \cite{cho2014learning} to transform $\mathbf{U}^t$ into a sequence of annotations $\mathbf{H}^t \in \mathbb{R}^{(N+2) \times 2d}$.

In practice, the pre and post-context vectors initialize the left-to-right and right-to-left RNNs with information about the utterances preceding and following $\mathbf{U}^t$. This is similar in spirit to the warm-start method of \citet{wang2017exploiting}, that directly initializes the hidden states of the RNNs with the context vectors. How we derive the pre and post-context vectors is explained in subsection \ref{subsec:context}.\\
\noindent \textbf{Self-attention}.
The self-attention mechanism \cite{vaswani2017attention,lin2017structured,yang2016hierarchical}, also called \textit{inner} or \textit{intra} attention, emerged in the literature following the success of attention in the sequence-to-sequence setting \cite{bahdanau2015neural,luong2015effective}. While self-attention deals with a single source sequence (no decoder), the motivation is the same as with traditional attention: rather than considering the last annotation of the RNN encoder as a summary of the entire input sequence, which is prone to information loss, a new hidden representation is computed as a weighted sum of the annotations at \textit{all} positions, where the weights are computed by a trainable mechanism that performs a comparison operation. 

While in seq2seq, the comparison involves the transformed input and the current hidden state of the decoder, in the encoder-only setting, the annotations $\mathbf{H}^t$ are passed through a dense layer and compared (dot product) with a trainable vector $\mathbf{u}_{\gamma}$, initialized randomly. Then, a probability distribution over the $N+2$ tokens in $\mathbf{U}^t$ is obtained via a softmax: 
{
\setlength{\abovedisplayskip}{3pt}
\setlength{\belowdisplayskip}{3pt}
\begin{equation}\label{eq:selfatt}
\pmb{\gamma}^t = \mathrm{softmax}(\mathbf{u}_{\gamma}\cdot \mathrm{tanh}(\mathbf{W}_{\gamma}\mathbf{H}^t))
\end{equation}
}
\noindent (bias omitted for readability). The attentional vector for $\mathbf{U}^t$ is finally computed as a weighted sum of its annotations, and, as shown in Fig. \ref{fig:architecture}, is finally passed to a dense layer to obtain the utterance embedding $\mathbf{u}^t \in \mathbb{R}^{d_f}$:

\begin{equation}
\mathbf{u}^t = \mathrm{dense}\Bigg(\sum_{i=1}^{N+2} \gamma^{t}_i \mathbf{h}_{i}^t\Bigg)
\end{equation}

\noindent $\mathbf{u}_{\gamma}$ replaces the hidden state of the decoder in the traditional attention mechanism. It can be interpreted as a learned representation of the ``ideal word'', on average. The more similar a token vector is to this representation, the more attention the model pays to the token.

\subsection{Context encoder: level 1}\label{subsec:context}
We now explain how we derive the pre and post-context vectors that we prepend and append to $\mathbf{U}^t$ so as to inject contextual information into the encoding process. They are obtained by aggregating information from the $C$ utterances preceding and following $\mathbf{U}^t$ (respectively):

{\small
\setlength{\abovedisplayskip}{0pt}
\setlength{\belowdisplayskip}{0pt}
\begin{align}
\mathbf{u}_{\mathrm{pre}}^t &\leftarrow \mathrm{aggregate_{pre}}\big( \big\{\mathbf{U}^{t-C},\dots,\mathbf{U}^{t-1}\big\}\big)\\
\mathbf{u}_{\mathrm{post}}^t &\leftarrow \mathrm{aggregate_{post}}\big( \big\{\mathbf{U}^{t+1},\dots,\mathbf{U}^{t+C}\big\}\big)
\end{align}
}

\noindent where $C$, the context size, is a hyperparameter. Since $\mathbf{u}_{\mathrm{pre}}^t$ and $\mathbf{u}_{\mathrm{post}}^t$ will become part of utterance $\mathbf{U}^t$ which is a sequence of token vectors, and fed to the RNN, we need them to live in the same space as any other token vector. This forbids the use of any nonlinear or dimension-changing transformation in $\mathrm{aggregate}$, such as convolutional or recurrent operations. Therefore, we use self-attention only. More precisely, we propose a two-level hierarchical architecture that makes use of a different type of self-attention at each level (see left part of Fig. \ref{fig:architecture}). The pre and post-context encoders share the exact same architecture, so we only describe the pre-context encoder in what follows.

\noindent \textbf{Content-aware self-attention}. At level 1, we apply the same attention mechanism to each utterance in $\big\{\mathbf{U}^{t-C},\dots,\mathbf{U}^{t-1}\big\}$. E.g., for $\mathbf{U}^{t-1}$:
{\small
\setlength{\abovedisplayskip}{0pt}
\setlength{\belowdisplayskip}{0pt}
\begin{equation}
\pmb{\alpha}^{t-1} = \mathrm{softmax}\bigg(\mathrm{\mathbf{u}_{\alpha}}\cdot \mathrm{tanh}\Big(\mathbf{W_{\alpha}}\mathbf{U}^{t-1} + \mathbf{W'}\sum_{i=1}^{N}{\mathbf{u}_{i}^{t}}\Big)\bigg)
\end{equation}
}
\noindent This mechanism is the same as in Eq. \ref{eq:selfatt}, except for two differences. First, we operate directly on the matrix of token vectors of the previous utterance $\mathbf{U}^{t-1}$ rather than on RNN annotations. Second, there is an extra input that consists of the element-wise sum of the token vectors of the current utterance $\mathbf{U}^{t}$.
The latter modification is inspired by the coverage vectors used in translation and summarization to address under(over)-translation and repetition, e.g., \cite{tu2016modeling,pointer-generator}.
In \cite{pointer-generator}, the coverage vector is the sum, over all previous steps of the decoder, of the attentional distributions over the source words.
Its role is to decrease repetition in the final summary, by letting the attention mechanism know which information about the source document has already been captured, in the hope that the model will focus on other aspects of it.
In our case, we hope that by letting the model know about the tokens in the current utterance $\mathbf{U}^{t}$, it will be able to extract complementary (rather than redundant) information from its context, and thus produce a richer embedding.

\noindent \textbf{Bi-directional information pathway}. To recapitulate, we consider $\mathbf{U}^{t}$ when computing $\mathbf{u}_{\mathrm{pre}}^t$ and $\mathbf{u}_{\mathrm{post}}^t$, and then prepend/append these vectors to $\mathbf{U}^{t}$ when encoding it. Therefore, in effect, information first flows from the current utterance to its context to guide context encoding, and then flows back to the current utterance encoding mechanism.

\noindent \textbf{Weight sharing}. The same content-aware self-attention mechanism is applied to the entire context surrounding $\mathbf{U}^{t}$, that is, to all preceding and following utterances. We did experiment with separate pre/post mechanisms, without significant improvements. This makes sense, as there is no inherent difference between preceding and following utterances. Indeed, the latter become the former as we slide the window over the transcription from start to finish. In addition, sharing weights makes for a more parsimonious and faster model. One should note, however, that the pre and post-context encoders still differ in terms of their time-aware attention mechanisms (at level 2).

\noindent \textbf{Dimensionality reduction}. The content-aware attention mechanism transforms the sequence of utterance matrices $\big\{\mathbf{U}^{t-C},\dots,\mathbf{U}^{t-1}\big\} \in \mathbb{R}^{C \times N \times d}$ into a sequence of vectors $\big\{\mathbf{u}^{t-C},\dots,\mathbf{u}^{t-1}\big\} \in \mathbb{R}^{C \times d}$. These vectors are then aggregated into a single pre-context vector $\mathbf{u}_{\mathrm{pre}}^t \in \mathbb{R}^{d}$ as described next.

\subsection{Context encoder: level 2}
As can be seen in Fig. \ref{fig:bipartite}, two utterances close to each other in time are much more likely to be related (e.g., adjacency pair, elaboration...) than any two randomly selected utterances. To enable our model to capture such time dependence, we used the trainable universal time-decay attention mechanism of \citet{su2018time}. 

\noindent \textbf{Time-aware self-attention}. The mechanism combines three types of time-decay functions via weights $w_i$. The attentional coefficient for $\mathbf{u}^{t-1}$ is:

{\small
\setlength{\abovedisplayskip}{0pt}
\setlength{\belowdisplayskip}{0pt}
\begin{align} \label{eq:timeatt}
\beta^{t-1} & = w_1 \beta^{\text{conv}^{t-1}} + w_2 \beta^{\text{lin}^{t-1}} + w_3 \beta^{\text{conc}^{t-1}}\\
 & = \frac{w_1}{a(d^{t-1})^b} + w_2[e d^{t-1}+k]^+ + \frac{w_3}{1+\big(\frac{d^{t-1}}{D_0}\big)^l}
\end{align}
}

\noindent where {\small$[\ast]^+$=$max(\ast, 0)$} (ReLU), $d^{t-1}$ is the offset between the positions of $\mathbf{U}^{t-1}$ and $\mathbf{U}^{t}$, i.e., {\small$d^{t-1}=|t-(t-1)|=1$}, and the $w_i$'s, $a$, $b$, $e$, $k$, $D_0$, and $l$ are scalar parameters learned during training. 

\noindent The convex (\texttt{conv}), linear (\texttt{lin}), and concave (\texttt{conc}) terms each model a different type of time dependence. Respectively, they assume the strength of dependence to weaken rapidly, linearly, and slowly, as the distance in time increases.
The post-context mechanism can be obtained by symmetry. It has different parameters.

\section{Community detection}
Once the utterance encoder $G_W$ presented in section \ref{sec:proposed} has been trained within the siamese or triplet meta-architecture presented in section \ref{sec:energy}, it is used to project the summary-worthy utterances from a given test transcription to a compact embedding space. We assume that if training was successful, the distance in that space encodes community structure, so that a basic clustering algorithm such as $k$-means \citep{macqueen1967} is enough to capture it. However, since we need to detect overlapping communities, we use a probabilistic version of $k$-means, the Fuzzy c-Means (FCM) algorithm \citep{bezdek1984fcm}. FCM returns a probability distribution over all communities for each utterance. More details are provided in appendix \ref{app:fcm}.

\section{Experiments}

\subsection{Dataset}
We experiment on the AMI corpus \citep{mccowan2005ami}, with the manual annotations v1.6.2. The corpus contains data for more than 100 meetings, in which participants play 4 roles within a design team whose task is to develop a prototype of TV remote control. Each meeting is associated with the annotations described in the introduction and shown in Fig. \ref{fig:bipartite}. There are 2368 unique abstractive communities in total, whose statistics are shown in Table \ref{table:communities}. We adopt the officially suggested \textit{scenario-only partition}\footnote{{\tiny\url{http://groups.inf.ed.ac.uk/ami/corpus/datasets.shtml}}}, which provides 97, 20, and 20 meetings respectively for training, validation and testing. We use manual transcriptions, and do not apply any particular preprocessing except filtering out specific ASR tags, such as \texttt{vocalsound}.

\begin{table}[H]
\setlength{\tabcolsep}{1.8pt}
\small
\centering
\scalebox{0.9}{
\begin{tabular}{r|cccc|c}
\hline
	 type & \texttt{abstract} & \texttt{action} & \texttt{problem} & \texttt{decision} & total \\ 
\hline
     unique     & 1147 & 247 & 380 & 594 & 2368 \\
\hline
     disjoint    & 528  & 124 & 69 & 45 & 766 \\
     nested      & 96 & 106 & 200 & 437 & 839 \\
     overlapping & 349 & 17 & 163 & 149 & 678\\
     singleton   & 49 & 162 & 38 & 244 & 493 \\
\hline
\end{tabular}
}
\caption{Statistics of abstractive communities.}
\label{table:communities}
\end{table}

\subsection{Baselines}
Full baseline details are provided in App \ref{app:baselines}.\\
$\bullet$ \noindent \textbf{Encoders}.
First, we evaluate our utterance encoder against two encoders that are trained within the energy framework: (1) \textbf{LD} \citep{lee2016sequential}, a sequential sentence encoder developed for dialogue act classification; and (2) \textbf{HAN} \citep{yang2016hierarchical}, a hierarchical self-attentive network for document embedding. We also compare our full pipeline against unsupervised and supervised systems. Note that to be fair, we ensure that both LD and HAN have access to context (see details in App B).

\noindent $\bullet$ \textbf{Unsupervised systems}.
In (1) \textbf{tf-idf}, we combine the TF-IDF vectors of the current utterance and the context utterances, each concatenated with their discourse features, and apply FCM.
In (2) \textbf{w2v}, we repeat the same approach with the word2vec centroids of the words in each utterance.
We also compare our full pipeline against \textbf{LCseg} \citep{galley2003discourse}, a lexical-cohesion based topic segmenter that directly clusters utterances without computing embeddings.

\noindent $\bullet$ \textbf{Supervised systems}. Finally, here, we use an approach similar to that of \citet{murray2012using}. More precisely, we train a MLP to learn abstractive links between utterances, and then apply the CONGA community detection algorithm to the utterance graph. 

We also considered 4 variants of our model:
(1) \textbf{CA-S}: we replace the time-aware self-attention mechanism of the context encoder with basic self-attention.
(2) \textbf{S-S}: we replace both the content-aware and the time-aware self-attention mechanisms of the context encoder with basic self-attention.
(3) \textbf{(0,0)}: our model, without using the contextual encoder.
(4) \textbf{(3,0)}: our model, using only pre-context, with a small window of 3, to enable fair comparison with the LD baseline.

\subsection{Training details}\label{sub:training}
\noindent \textbf{Word encoder}. Discourse features consist of two one-hot vectors of dimensions 4 and 16, respectively for speaker role and dialogue act. The positional feature is a scalar in $[0,1]$, indicating the normalized position of the utterance in the transcription. We used the pre-trained vectors learned on the Google News corpus with word2vec by \citep{mikolov2013exploiting}, and randomly initialized out-of-vocabulary words (1645 out of 12412). As a preprocessing step, we reduced the dimensionality of the pre-trained word vectors from 300 to 21 with PCA, in order to give equal importance to discourse and textual features. In the end, tokens are thus represented by a $d=42$-dimensional vector.

\noindent \textbf{Layer sizes}.
For our model, and the LD and HAN baselines, we set $d_f=32$ (output dimension of the final dense layer).

\noindent \textbf{LD}. We set d1=3 and d2=0, which is very close to (2,0), the best configuration reported in the original paper.

\noindent \textbf{HAN}. Again, for the sake of fairness, we give the HAN baseline access to contextual information, by feeding it the current utterance surrounded by the $C_b$ preceding and $C_b$ following utterances in the transcription, where $C_b$ denotes the best context size reported in section \ref{sec:res}.

\noindent \textbf{Training details}. The exact same token representations and settings were used for our model, its variants, and the baselines. Models were trained on the training set for 30 epochs with the Adam \cite{kingma2014adam} optimizer.
The best epoch was selected as the one associated with the lowest validation loss. Batch size and dropout \citep{srivastava2014dropout} were set to 16 and 0.5. Dropout was applied to the word embedding layer only. To account for randomness, we average results over 10 runs.
Also, following \citep{hoffer2015deep,liu2019dynamic}, we use a different, small subset of all possible triplets for training at each epoch (more precisely, 15594 triplets). This intelligently maximizes data usage while preventing overfitting.
To enable fair comparison with the siamese approach, 15594 genuine and 15594 impostor pairs were sampled at the beginning of each epoch, since we consider that one triplet essentially equates one genuine pair and one impostor pair.

\noindent \textbf{Performance evaluation}.
We evaluate performance at the distance and the clustering level, using respectively precision, recall, and F1 score at $k$, and the omega index \citep{collins1988omega}. For P, R, and F1, we evaluate the quality of the ranking of the closest utterances to a given query utterance. We use a fixed $k$=10 and also a variable $k$ (denoted as $k$=v), where $k$ is equal to the size of the community of the query utterance minus one. In that case, P=R=F1. More details and examples are given in appendices \ref{app:ranking_example} and \ref{app:perf}.

For the omega index, we report results with a fixed number of communities $|Q|$=11, and also a variable $|Q|$ ($|Q|$=v), where $|Q|$ is equal to the number of ground truth communities. More details and examples are given in App \ref{app:perf}.

Due to the stochastic nature of the FCM algorithm, we select the run yielding the smallest objective function value over 20 runs.

\section{Results}\label{sec:res}
\begin{figure}[ht]
    \centering
    \includegraphics[scale=0.45]{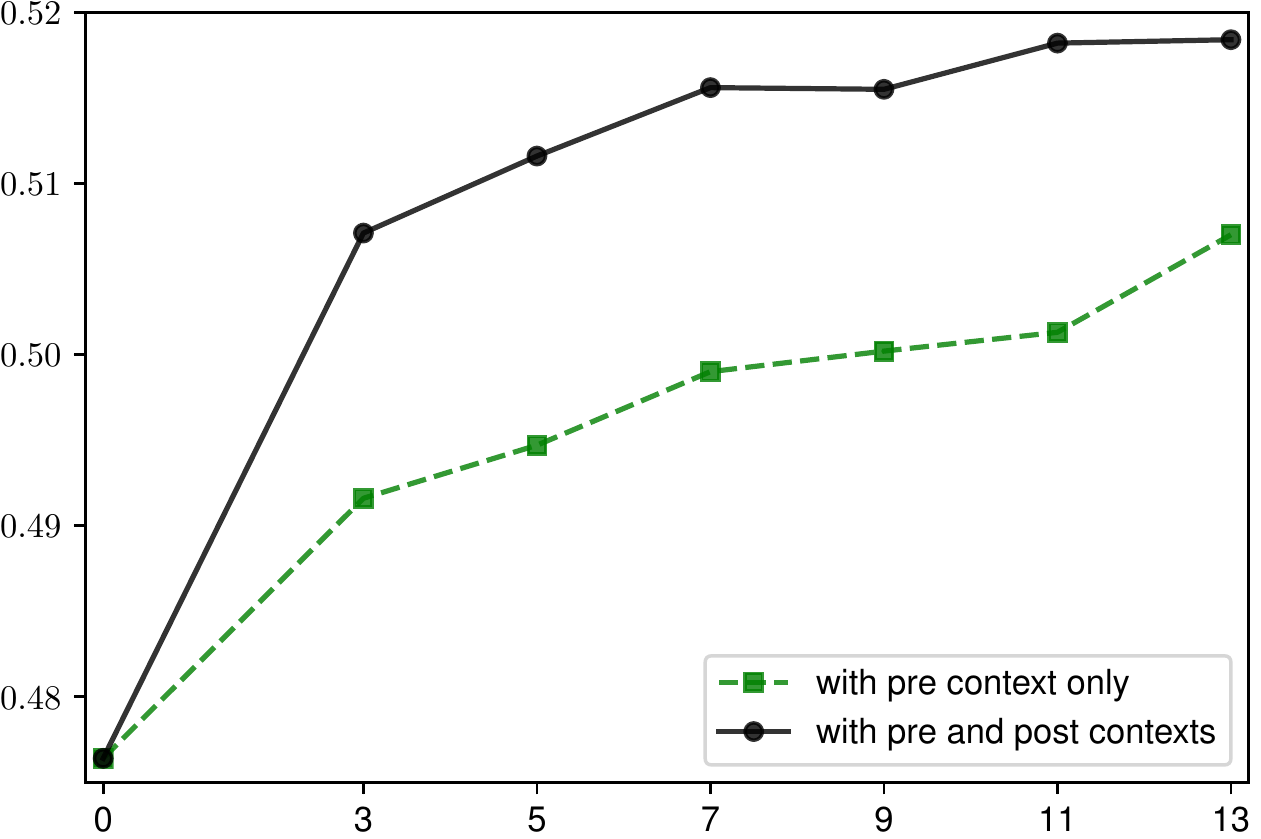}
	\caption{Impact of context size on the validation $P@k=v$, for our model trained within the triplet meta-architecture.
	}\label{fig:val_prf1_kv_vs_context_size}
\end{figure}

\noindent \textbf{Context sizes}
Larger contexts bring richer information, but increase the risk of considering unrelated utterances.
Using our proposed encoder within the triplet meta-architecture, we tried different values of $C$ on the validation set, under two settings: $(\mathrm{pre},\mathrm{post})=(C,0)$, and $(\mathrm{pre},\mathrm{post})=(C,C)$. Results are shown in Fig. \ref{fig:val_prf1_kv_vs_context_size}. We can observe that increasing $C$ always brings improvement, with diminishing returns. Results also clearly show that considering the following utterances in addition to the preceding ones is useful. Note that the curves look similar for $F1@k=10$. In the end, we selected (11,11) as our best context sizes.

\begin{table*}[ht]
\small
\centering
\setlength{\tabcolsep}{6pt}
\renewcommand{\arraystretch}{1.1}
\scalebox{0.97}{
\begin{tabular}{|rrrr||c|ccc||c|c|}
  \hline
  \multirow{2}{*}{}
    & \multicolumn{1}{c}{}
    & \multicolumn{1}{c}{}
    & \multicolumn{1}{c||}{(pre,}
    & \multicolumn{1}{c|}{P}
    & \multicolumn{1}{c}{P} 
    & \multicolumn{1}{c}{R}
    & \multicolumn{1}{c||}{F1} 
    & \multicolumn{2}{c|}{{\small Omega index $\times 100$}} 
    \\  \cline{5-10}
    & \multicolumn{1}{c}{}
    & \multicolumn{1}{c}{}
    & \multicolumn{1}{c||}{post)}
    & \multicolumn{1}{c|}{$@k=v$} 
    & \multicolumn{3}{c||}{$@k=10$} 
    & \multicolumn{1}{c|}{{\tiny$|Q|=v$}}
    & \multicolumn{1}{c|}{{\tiny$|Q|=11$}} 
    \\  \cline{1-10}
    
             & \texttt{a1}) & our model                        & (0, 0)   & 54.59 & 46.05 & 62.45 & 43.18 & 49.09 & 48.81 \\
             & \texttt{a2}) & our model                        & (3, 0)   & 55.17 & 46.17 & 62.80 & 43.25 & 49.78 & 49.70 \\
      & \texttt{a3}) & our model                        & (11, 11) & 58.58 & 46.73 & 63.82 & 43.83 & 49.90 & 49.28 \\
Triplet & \texttt{b}) & our model (CA-S) & (11, 11) & \textbf{59.52}$^\star$ & \textbf{46.98}$^\star$ & \textbf{64.01}$^\star$ & \textbf{44.06}$^\star$ & \textbf{50.11} & 49.73 \\
             & \texttt{c}) & our model (S-S)          & (11, 11) & 58.96 & 46.81 & 63.65 & 43.87 & 49.59 & \textbf{49.88} \\
\cline{3-10}
             & \texttt{d}) & LD        & (3, 0)   & 52.04 & 44.82 & 60.41 & 41.82 & 48.70 & 48.14 \\
             & \texttt{e}) & HAN                              & (11, 11) & 58.72 & 45.76 & 62.60 & 42.89 & 49.32 & 48.88 \\
\hline
\hline
             & \texttt{f1}) & our model                        & (0, 0)   & 53.01 & 45.10 & 60.97 & 42.12 & 50.56 & 49.65 \\
             & \texttt{f2}) & our model                        & (3, 0)   & 53.78 & 45.54 & 61.33 & 42.48 & 51.01 & 50.00 \\
      & \texttt{f3}) & our model                        & (11, 11) & 56.64 & \textbf{46.47} & \textbf{62.54} & \textbf{43.40} & \textbf{52.44}$^\star$ & \textbf{51.88}$^\star$ \\
Siamese & \texttt{g}) & our model (CA-S) & (11, 11) & 56.46 & 46.08 & 61.92 & 43.02 & 51.60 & 50.98 \\
             & \texttt{h}) & our model (S-S)         & (11, 11) & 55.68 & 45.64 & 61.17 & 42.53 & 52.26 & 51.11 \\
\cline{3-10}
             & \texttt{i}) & LD & (3, 0)   & 52.13 & 44.83 & 60.85 & 41.86 & 51.18 & 50.70  \\
             & \texttt{j}) & HAN                              & (11, 11) & \textbf{58.54} & 45.72 & 61.55 & 42.74 & 50.51 & 49.82  \\
\hline
\hline
             & \texttt{k1}) & tf-idf                           & (0, 0)   & 29.28 & 26.67 & 34.69 & 24.19 & 13.12 & 13.66 \\
             & \texttt{k2}) & tf-idf                           & (3, 0)   & 34.77 & 30.27 & 40.83 & 27.79 & 10.22 & 10.17 \\
             & \texttt{k3}) & tf-idf                           & (11, 11)   & \textbf{58.94} & 43.94 & 61.36 & 41.45 & 38.09 & 39.47 \\
           Unsupervised & \texttt{l1}) & w2v                         & (0, 0)   & 29.02 & 27.46 & 37.39 & 25.11 & 13.89 & 13.50  \\
             & \texttt{l2}) & w2v                         & (3, 0)   & 34.11 & 29.92 & 39.55 & 27.32 & 10.61 & 10.77 \\
             & \texttt{l3}) & w2v                         & (11, 11)   & 58.30 & \textbf{44.08} & \textbf{61.59} & \textbf{41.59} & 37.75 & 38.28 \\
             & \texttt{m}) & LCSeg                            & - & - & - & - & - & \textbf{38.98} & \textbf{41.57} \\
\hline
\hline
             & \texttt{n1}) &  tf-idf                 & (0, 0)   & - & - & - & - & 25.04 & 25.14 \\
             & \texttt{n2}) &  tf-idf                 & (3, 0)   & - & - & - & - & 27.33 & 26.95 \\
            Supervised & \texttt{n3}) &  tf-idf           & (11, 11)   & - & - & - & - & \textbf{45.26} & \textbf{44.91} \\
              & \texttt{o1}) &  w2v             & (0, 0)   & - & - & - & - & 25.32 & 25.25 \\
              & \texttt{o2}) &  w2v               & (3, 0)   & - & - & - & - & 29.14 & 29.02 \\
              & \texttt{o3}) &  w2v                & (11, 11)   & - & - & - & - & 43.31 & 43.08 \\
\hline
\end{tabular}
}
\caption{Results (averaged over 10 runs). $^\star$: best score per column. \textbf{Bold}: best score per section. -: does not apply as the method does not produce utterance embeddings. \label{table:results}}
\end{table*}

\noindent \textbf{Quantitative results}.
\noindent Final test set results are shown in Table \ref{table:results}. All variants of our model significantly outperform LD. While HAN is much stronger than LD, our model and its variants using best context sizes manage to outperform it everywhere, except in the siamese/P@k=v case (row j).
One of the reasons for the superiority of our utterance encoder is probably that it considers contextual information \textit{while encoding} the current utterance, while HAN and LD take as input the context utterances together with the current utterance, without distinguishing between them. Moreover, we use an attention mechanism dedicated to temporality, whereas HAN is only able to capture an implicit notion of time through the use of recurrence (RNN), and LD, with its dense layers, completely ignores it.
Also, all variants of our model using best context sizes (11,11) outperform the ones using reduced (3,0) or no (0,0) context, regardless of the meta-architecture. This confirms the value added by our context encoder.

For siamese, our model outperforms its two variants (\textit{CA-S} and \textit{S-S}) for all metrics, indicating that both the content-aware and the time-aware self-attention mechanisms are useful.
However, it is interesting to note that when training under the triplet configuration, the CA-S variant of our model is better, suggesting that in that case, the content-aware mechanism is beneficial, but the time-aware one is not.

LCseg (row m) and tf-idf (11,11) (row n3) are the best of all (un)supervised baseline systems, but both perform significantly worse than all energy-based approaches, highlighting that training with the energy framework is beneficial. In terms of Omega Index, supervised baseline systems are logically better than unsupervised ones.

w2v generally outperforms tf-idf when there is no context (rows k1,l1,n1,o1) or short context (k2,l2,n2,o2), but not with large contexts (k3,l3,n3,o3).
Results also show that overall, using larger contexts always brings improvement.

\noindent \textbf{Qualitative results}.
We visualize in App \ref{app:vis_attentions} that the three self-attention mechanisms behave in a cooperative manner to produce a meaningful utterance representation. We also visualize the attention coefficients of the two time-aware self-attention mechanisms, and find that interestingly, the distributions over the pre and post-context are not symmetric. We also inspect the closest utterances to a given query utterance in App \ref{app:ranking_example}.

\noindent \textbf{Simplified task}.
Finally, we also experimented on a much simpler task, where only the communities of type \texttt{ABSTRACT} were considered. This makes ACD much simpler, because most of the overlapping communities are of the other types (see Table \ref{table:communities}).
For this simplified task, we have 1147 unique communities, of which 78.99\% are disjoint. our model achieves 72.09 in terms of $P@k=v$ and 55.67 in terms of Omega Index when $|Q|=v$. $P,R,F1@k=15$ are respectively equal to 55.07, 74.37, and 54.00, and the Omega Index is 54.30 when $|Q|=8$.

\section{Conclusion}
This paper proposes one of the first applications of energy-based learning to ACD. Using the siamese and triplet meta-architectures, we showed that our novel contextual utterance encoder learns better distance and communities than state-of-the-art competitors.

\section*{Acknowledgments}
This research was supported in part by the OpenPaaS::NG and LinTo projects.

\bibliography{main}
\bibliographystyle{acl_natbib}

\newpage
\onecolumn

\begin{center}
\textbf{\Large Energy-based Self-attentive Learning of Abstractive Communities for Spoken Language Understanding}
\vspace{0.2cm}

\textbf{\Large \textit{Supplementary Material}}
\vspace{0.5cm}
\end{center}

\noindent\textbf{\Large Appendices}
\appendix

\section{Attention visualization}\label{app:vis_attentions}
\begin{figure}[H]
\centering
\includegraphics[scale=0.6]{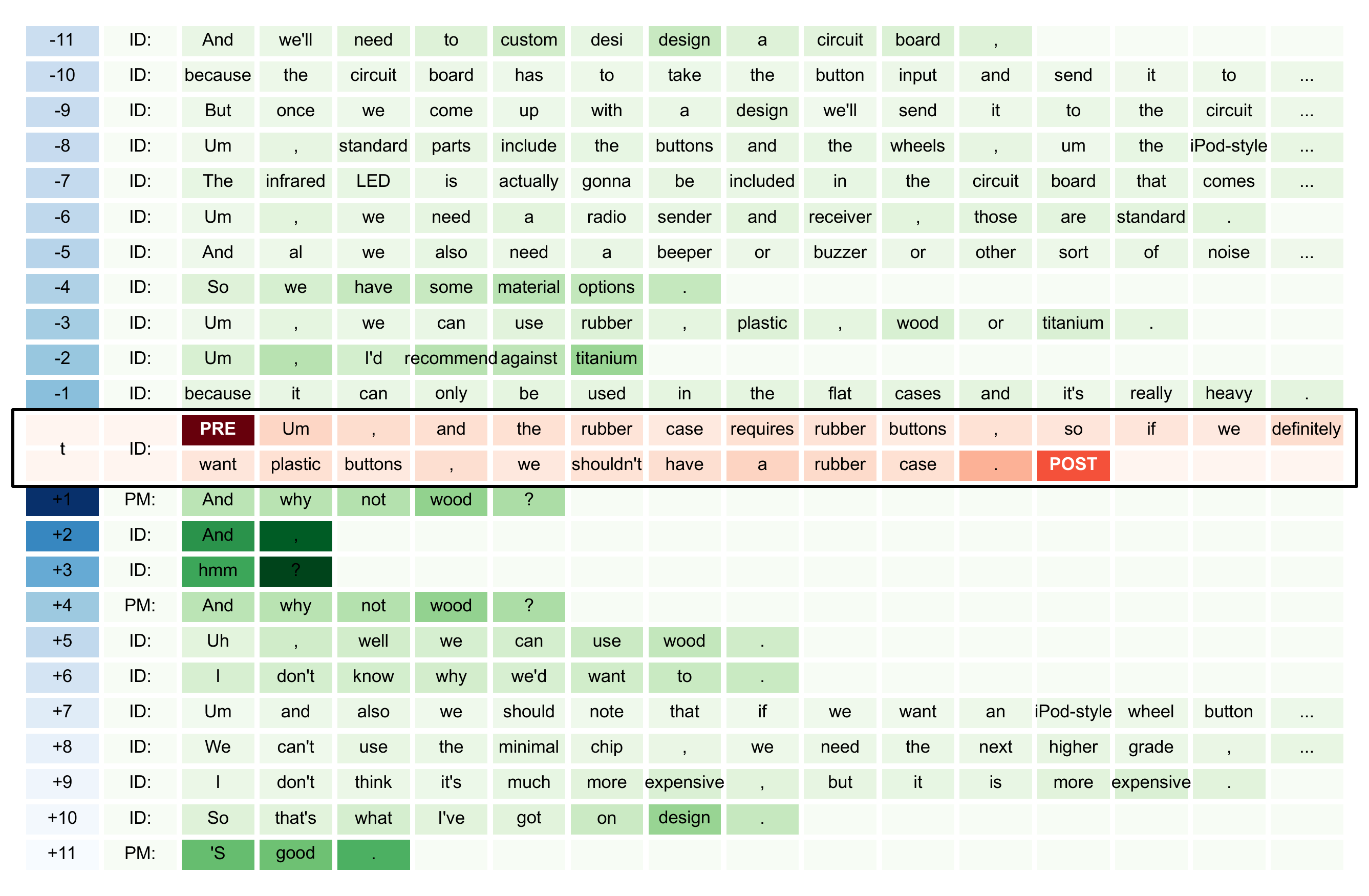}
\caption{Visualization of attention distributions around an utterance from the ES2011c meeting. Some utterances are truncated for readability.}
\label{fig:vis_attentions}
\end{figure}

The aim of this section is to show, with an example, what the three self-attention mechanisms pay attention to while encoding the current utterance $\mathbf{U}^{t}$ (here, an utterance from the ES2011c \textit{validation} meeting). Fig. \ref{fig:vis_attentions} shows the attention distributions over $\mathbf{U}^t$ (highlighted by the black frame), and over its pre-context $\{ \mathbf{U}^{t-1},\ldots,\mathbf{U}^{t-11} \}$ and post-context $\{ \mathbf{U}^{t+1},\ldots,\mathbf{U}^{t+11} \}$ utterances. We use three colors that are consistent with the ones used in Fig. \ref{fig:architecture} to denote the three different attention mechanisms: green for content-aware ($\alpha$), blue for time-aware ($\beta$), and red for basic self-attention ($\gamma$). Remember that $\alpha$ and $\beta$ are both in the context encoder, while $\gamma$ is in the utterance encoder. Color shades indicate attention intensity (the darker, the stronger).

\noindent We can observe in Fig. \ref{fig:vis_attentions} that:
\begin{itemize}
    \item The content-aware self-attention mechanism $\alpha$ (green) focuses on the informative and complementary words in the contexts that are central to understanding the utterance at time $t$, such as: ``custom", ``design" from $\mathbf{U}^{t-11}$, ``material" from $\mathbf{U}^{t-4}$, ``recommend", ``titanium" from $\mathbf{U}^{t-2}$, ``wood" from $\mathbf{U}^{t+1}$, etc.
    \item The time-aware self-attention mechanism $\beta$ (blue) places more importance over the context utterances that are close to $\mathbf{U}^t$, i.e., the importance decreases when the time distance increases. However, the patterns are different for the pre and post-contexts (see Fig. \ref{fig:timeatt} below).
    \item The self-attention mechanism $\gamma$ (red) focuses mainly on the special pre-context token \texttt{PRE}, meaning that the pre-context is more important than the post-context in the example considered. Generally speaking, the pre and post-context tokens contain richer information than any token from the current utterance, as the context tokens originate from the fusion of $\{ \mathbf{U}^{t-11},\ldots,\mathbf{U}^{t},\ldots,\mathbf{U}^{t+11} \}$. It is thus possible that the utterance encoder has learned to always pay more attention to these information-rich tokens than to any regular token.
    \item It is also interesting to note that considerable attention is being paid to punctuation marks. This makes sense, since they are important pieces of information indicative of utterance type (e.g., statement or question).
\end{itemize}

To summarize, the visualization results show that the three self-attention mechanisms of our model are able to adaptively focus on different information, in order to cooperatively produce a meaningful representation.\\

\noindent We also inspect in Fig. \ref{fig:timeatt} the attention coefficients of the time-aware self-attention mechanisms (see Eq. \ref{eq:timeatt}) equipping the pre and post-context encoders. It is interesting to observe that the distributions are not symmetric. Indeed, only the utterances immediately following $\mathbf{U}^{t}$ ($t+1 \rightarrow t+5$) seem to matter, while the attention weights are much more uniform across the utterances preceding $\mathbf{U}^{t}$. This suggests that in dialogues, considering a long history of preceding utterances helps understanding the current one.
\begin{figure}[ht]
    \centering
    \includegraphics[scale=0.7]{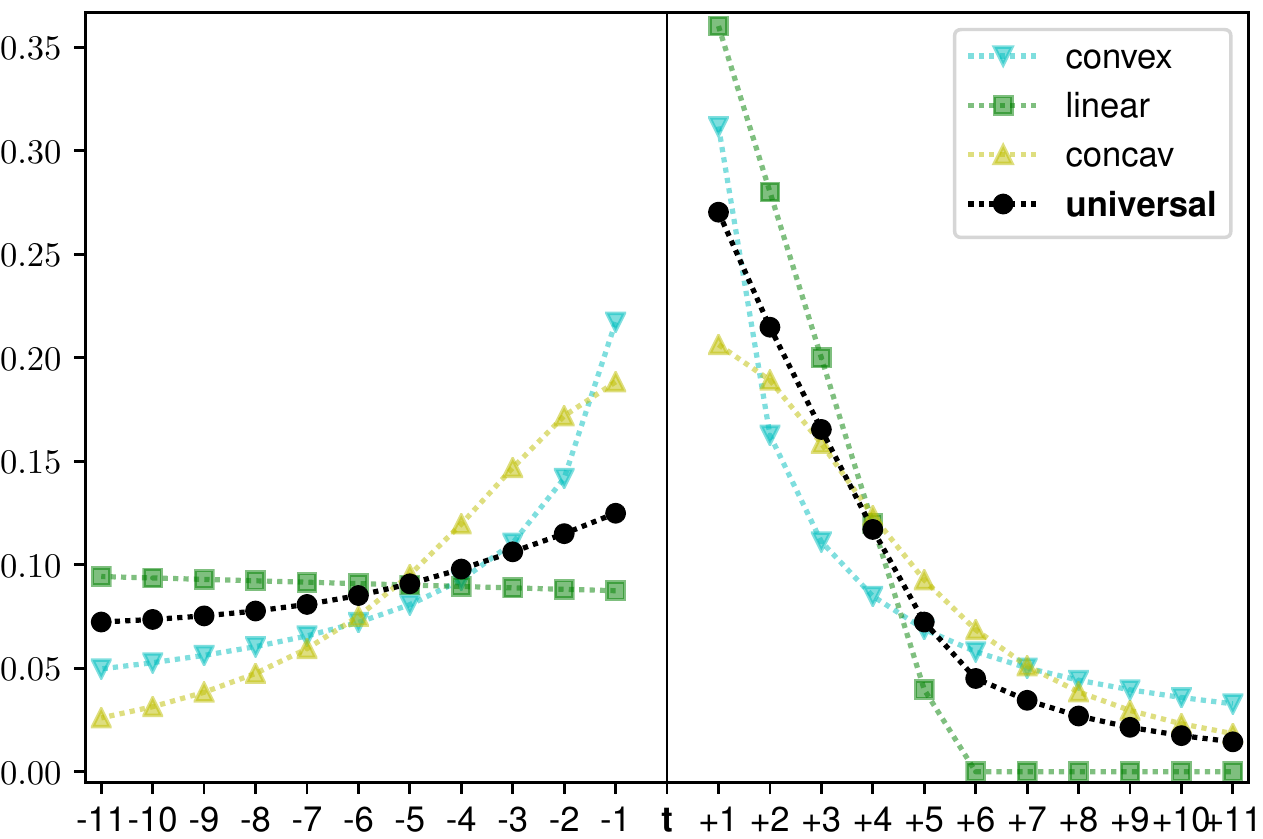}
	\caption{Normalized time-aware self-attention weights for pre and post-contexts, averaged over 10 runs.
	}   \label{fig:timeatt}
\end{figure}

\noindent It is also interesting to note that the parameters that have been learned for the pre-context linear function make it increasing, rather than decreasing. This is counter-intuitive, but allowed by design. Overall though, the three terms altogether do produce a function that slowly decreases as time distance increases, which is in accordance with intuition.

\section{Baselines}\label{app:baselines}
\subsection{Baseline encoders}

\noindent $\bullet$ \textbf{LD} \citep{lee2016sequential} is a sequential sentence encoder developed for dialogue act classification. The model takes into account a fixed number of utterances from the pre-context when classifying the current one. More precisely, CNN or RNN with max-pooling is first applied separately to the current utterance and each pre-context utterance, and the resulting vectors are then aggregated through two levels of dense layers, based on two hyper-parameters, $d1$ and $d2$, which represent the history size at level 1 and level 2 (respectively).
Although the original paper reported that the CNN encoder slightly outperforms the RNN one (for DA classification), in our experiments, we used the RNN variant, since our model and the HAN baseline are RNN-based. Note that here, we used LSTM cells as \citet{lee2016sequential} reported them to work better than GRU cells in their experiments.

\noindent $\bullet$ \textbf{HAN} \citep{yang2016hierarchical}. The Hierarchical Attention Network, developed for document classification, is a two-level architecture, where at level 1, each sentence in the document is separately encoded by the same sentence encoder, resulting in a sequence of sentence vectors. 
That sequence is then processed at level 2 by the document encoder which returns a single vector representing the entire document. 
The sentence and document encoders are both self-attentional bidirectional Recurrent Neural Networks (RNNs), with different parameters.
We give HAN access to contextual information by feeding it the current utterance surrounded by the $C_b$ preceding and $C_b$ following utterances in the transcription, where $C_b$ denotes the best context size reported in section \ref{sec:res}.

\subsection{Unsupervised baseline systems}

\noindent $\bullet$ \textbf{tf-idf}. A TF-IDF vector is used as the utterance embedding, compressed to a dimension of 21 with PCA, and concatenated with the 21-dimensional discourse feature vector, thus forming a vector of dimension $d=42$. This vector is then again compressed to a $d_f=32$-dimensional vector.
The compression steps are applied for consistency with the energy-based systems, in which textual and discourse features have the same dimensionality $\nicefrac{d}{2}=21$, and the output of the utterance encoder is $d_f$-dimensional (see subsection 6.3).
To make this baseline context-aware, the embeddings of the current utterance and the context utterances are averaged. In the end, FCM is applied.
Note that the TF-IDF vocabulary is obtained from the entire conversation, giving this baseline a competitive advantage over the others, which never have access to the full transcription.

\noindent $\bullet$ \textbf{w2v}. Identical to the previous baseline, but using the average of the word2vec vectors of a given utterance instead of TF-IDF vector.

\noindent $\bullet$ \textbf{LCseg} is an unsupervised system adapted from previous work \citep{oya2014template,banerjee2015generating,singla2017automatic}, in which disjoint topic segments are assumed to be abstractive communities. A lexical-cohesion based topic segmenter LCseg \citep{galley2003discourse} is first applied on transcriptions to get the desired number of segments ($|Q|=v/11$), and then only summary-worthy utterances within segments are retained for evaluation.

\subsection{Supervised baseline systems}
As discussed in the literature review (see section 2), original approaches to ACD \citep{murray2012using,mehdad2013abstractive} are supervised and non energy-based. They have no publicly available implementations, and are hard to precisely reimplement due to lack of details about handcrafted features and dependency on external textual entailment corpora.
Nevertheless, we implemented two baselines similar in spirit, taking as input the representations produced by the tf-idf and w2v unsupervised baselines previously described.
More precisely, the two $d_f$-dimensional representations of a pair of utterances are fed into a 3-layer feed-forward neural network (with $2d_f$, $d_f$, and $1$ hidden units) which is trained on the task of predicting whether the two utterances belong to the same abstractive community or not (binary classification task).
Then, like in the aforelisted studies, an utterance graph is built, where utterances are linked based on the predictions of the MLP.
Finally, the CONGA algorithm \citep{gregory2007algorithm}, an extension of the well-known Girvan-Newman algorithm, is applied to detect overlapping communities on the utterance graph.

\section{Ranking example} \label{app:ranking_example}
For the same utterance from the ES2011c meeting as used in appendix \ref{app:vis_attentions}, we show below the closest and furthest utterances, in terms of Euclidean distance in the embedding space. Recall that meeting ES2011c belongs to the \textit{validation} set. Utterances belonging to the ground truth community of the query utterance are shown in bold. Roles are ID: industrial designer, ME: marketing expert, UI: user interface designer, PM: project manager. For this example, $P@k=v$ is equal to 77.78 (where $v=9$), and $P$, $R$, and $F1@k$ are 80.00, 88.89, 84.21 respectively (where  $k=10$).

We can see that semantic similarity obviously plays a role, as most of the closest utterances are about buttons and materials. But other parameters come into play. E.g., the utterances {\small\texttt{And al we also need a beeper or buzzer or other sort of noise thing for locating the remote}}, and {\small \texttt{I don't know why we'd want to}}, respectively ranked 2$^{\texttt{nd}}$ and 7$^{\texttt{th}}$, are not semantically related to the query utterance. Such utterances might be placed close to the query utterance based on their positional and discourse features (speaker role and dialogue act), but also because their contexts are similar.

\begin{table}[H]
\addtocounter{table}{-1}
\small
\setlength{\tabcolsep}{3pt}
\renewcommand{\arraystretch}{1.2}
\begin{tabularx}{\linewidth}{|r|rrrX|}
\hline
dist & pos & DA & role & text \\
\hline
0   & $t$   & inf    & ID & \textbf{Um , and the rubber case requires rubber buttons , so if we definitely want plastic buttons , we shouldn't have a rubber case .}                                      \\
\hline
0.11  & -3   & inf    & ID & Um , we can use rubber , plastic , wood or titanium .                                                                                                                \\
0.12  & -5   & inf    & ID & \textbf{And al we also need a beeper or buzzer or other sort of noise thing for locating the remote .}                                                                        \\
0.38  & -2   & sug    & ID & \textbf{Um , I'd recommend against titanium}                                                                                                                                  \\
0.42  & +7   & inf    & ID & Um and also we should note that if we want an iPod-style wheel button , it's gonna require a m qu slightly more expensive chip .                                     \\
0.54  & +5   & ass    & ID & \textbf{Uh , well we can use wood .}                                                                                                                                          \\
0.57  & -8   & inf    & ID & \textbf{Um , standard parts include the buttons and the wheels , um the iPod-style wheel .}                                                                                   \\
0.68  & +6   & ass    & ID & \textbf{I don't know why we'd want to .}                                                                                                                                      \\
0.96  & -11  & inf    & ID & \textbf{And we'll need to custom desi design a circuit board ,}                                                                                                               \\
1.26  & -13  & inf    & ID & \textbf{Um , I assume we'll be custom designing our case ,}                                                                                                                   \\
1.27  & -14  & inf    & ID & \textbf{Um , so we need some custom design parts , and other parts we'll just use standard .}                                                                                 \\
1.43  & -17  & inf    & ID & \textbf{So I've been looking at the components design .}                                                                                                                      \\
1.66  & +12  & off    & ME & Um , can I do next ? Because I have to say something about the material                                                                                              \\
2.24  & +18  & inf    & ME & and the findings are that the first thing to aim for is a fashion uh , fancy look and feel .                                                                         \\
2.57  & +19  & inf    & ME & Um . Next comes technologic technology and the innovations to do with that .                                                                                         \\
3.21  & +20  & inf    & ME & And th last thing is the easy to use um factor .                                                                                                                     \\
3.92  & +69  & inf    & UI & Uh , so people are going to be looking at this little screen .                                                                                                       \\
4.02  & +92  & inf    & ME & But the screen can come up on the telly , the she said .                                                                                                             \\
$\boldsymbol{\dotsm}$ & & & & \\
8.81  & +623 & inf    & ID & It didn't give me any actual cost .                                                                                                                                  \\
8.84  & +622 & inf    & ID & All it said was it gave sort of relative , some chips are more expensive than others , sort of things .                                                              \\
8.89  & +616 & inf    & ME & So if you throw it , it's gonna store loads of energy , and you don't need to buy a battery because they're quite f I find them annoying .                           \\
9.00   & +617 & sug    & ME & But we need to find cost .                                                                                                                                           \\
9.06  & +621 & el.inf & ME & Does anyone have costs on the on the web ?                                                                                                                           \\
9.95  & +652 & inf    & PM & And you're gonna be doing protu product evaluation .                                                                                                                 \\
9.96  & +650 & inf    & PM & Oh when we move on , you two are going to be playing with play-dough .                                                                                               \\
10.15 & +651 & inf    & PM & Um , and working on the look and feel of the design and user interface design .                                                                                      \\
\hline
\end{tabularx}
\end{table}

\noindent The community where the query utterance belongs to (utterances shown in bold in the table above) is associated with the following sentence in the human abstractive summary: \texttt{The Industrial Designer gave her presentation on components and discussed which would have to be custom-made and which were standard}.

\section{Performance evaluation}\label{app:perf}
We evaluate performance at the distance and the clustering level.

\subsection{Distance}
First, we test whether the distance in the final embedding space is meaningful. To do so, for a given \textit{query} utterance, we rank all other utterances in decreasing order of similarity with the query. We then use precision, recall, and F1 score at $k$ to evaluate the quality of the ranking. A detailed example was provided in App \ref{app:ranking_example}.

Singleton communities are excluded from the evaluation at this stage. We set $k$=10, which is equal to the average number of non-singleton communities minus one (since the query utterance cannot be part of the results). We also report results for a variable $k$ ($k$=v), where $k$ is equal to the size of the community of the query utterance minus one. In that case, P=R=F1.

The same procedure is repeated for all utterances. To account for differences in community size, scores are first averaged at the community-level, and then at the meeting-level. Note that the distance is Euclidean for triplet and Manhattan for siamese (see subsections \ref{sub:siam} and \ref{sub:trip}).

\subsection{Clustering}
Second, we compare our community assignments to the human ground truth using the Omega-Index
\citep{collins1988omega}, a standard metric for comparing non-disjoint clustering, used in the ACD literature \citep{murray2012using}.

The Omega Index evaluates the degree of \textit{agreement} between two clustering solutions based on \textit{pairs} of objects being clustered. Two solutions $s_1$ and $s_2$ are considered to agree on a given pair of objects, if two objects are placed by both solutions in \textit{exactly} the same \textit{number of communities} (possibly zero).

The Omega Index $\omega$ is computed as shown in Equation \ref{eq:omega}. The numerator is the observed agreement $\omega_{obs}$ adjusted by expected (chance) agreement $\omega_{exp}$, while the denominator is the perfect agreement (value equals to 1) adjusted by expected agreement.
\begin{equation} \label{eq:omega}
    \omega(s_1, s_2) = \frac{\omega_{obs}(s_1, s_2)-\omega_{exp}(s_1, s_2)}{1-\omega_{exp}(s_1, s_2)} \\
\end{equation}

Observed and expected agreements are calculated as below:
\begin{align}
    \omega_{obs}(s_1, s_2) &= \frac{1}{N_{total}} \sum_{j = 0}^{min(J, K)} A_j   \\
    \omega_{exp}(s_1, s_2) &= \frac{1}{N_{total}^2} \sum_{j = 0}^{min(J, K)} N_{j1}N_{j2}
\end{align}
where $A_j$ is the number of pairs agreed to be assigned to $j$ number of communities by both solutions,
$N_{j1}$ is the number of pairs assigned to $j$ communities in $s_1$,
$N_{j2}$ is the number of pairs assigned to $j$ communities in $s_2$,
$J$ and $K$ represent respectively the maximum number of communities in which any pair of objects appear together in solutions $s_1$ and $s_2$, and
$N_{total}=n(n-1)/2$ is the total number of pairs constructed over $n$ number of objects.

To give an example, consider two clustering solutions for 5 objects: 
\begin{align*}
    s_1&=\{ \{a,b,c\}, \{b,c,d\}, \{c,d,e\}, \{c,d\} \} \\
    s_2&=\{ \{a,b,c,d\}, \{b,c,d,e\} \}
\end{align*}

\begin{table}[H]
\addtocounter{table}{-1}
\setlength{\tabcolsep}{1.8pt}
\small
\centering
\scalebox{0.95}{
\begin{tabular}{r|c|c|c}
\hline
& solution $s_1$ & solution $s_2$ & solutions \\
\cline{2-3}
& \#communities & \#communities & $s_1$ and $s_2$\\
& the pair is assigned & the pair is assigned & agree on the pair? \\
\hline
(a, b) & 1 & 1 & yes \\
(a, c) & 1 & 1 & yes \\
(a, d) & 0 & 1 & no \\
(a, e) & 0 & 0 & yes \\
(b, c) & 2 & 2 & yes \\
(b, d) & 1 & 2 & no \\
(b, e) & 0 & 1 & no \\
(c, d) & 3 & 2 & no \\
(c, e) & 1 & 1 & yes \\
(d, e) & 1 & 1 & yes \\
\hline
\end{tabular}
}
\captionsetup{size=small}
\end{table}
Solutions are transformed into the table above, from what we can obtain $N_{total}=10, J=3, K=2, min(J,K)=2$. Two solutions agree to place $(a,e)$  together in no community, the pairs $(a,b)$, $(a,c)$, $(c,e)$ and $(d,e)$ in one community, and the pair $(b,c)$ in two communities. We have $A_0=1, A_1=4, A_2=1$. Thus the observed agreement is $(1+4+1)/10=0.6$.
Since
$N_{01}=3, N_{11}=5, N_{21}=1$ and 
$N_{02}=1, N_{12}=6, N_{22}=3$, the expected agreement then is $(3*1+5*6+1*3)/10^2=0.36$. Finally, Omega Index for this simple example is computed as: $\omega(s_1, s_2)=(0.6-0.36)/(1-0.36)=0.375$.

Since FCM yields a probability distribution over communities for each utterance, we need to use a threshold to assign a given utterance to one or more communities. We selected $0.2$ after trying multiple values in $[0, 0.5]$ with steps of $0.05$ on the validation set. Whenever one or more utterances were not assigned to any community, we merged them into a new community. Furthermore, we set the number of clusters $|Q|$ to 11, which corresponds to the average number of ground truth communities per meeting (after merging). We also report results with a variable $|Q|$ ($|Q|=v$), equal to the number of ground truth communities.

Note that since FCM does not return nested groupings, we merged the ground truth communities nested under the same community.

\section{FCM algorithm}\label{app:fcm}

The goal of the Fuzzy c-Means algorithm or FCM \citep{bezdek1984fcm} is to minimize the weighted within group sum of squared error objective function:
\begin{equation}
    J(M,Q)=\sum_{q=1}^{|Q|}\sum_{t=1}^{T} (m_{qt})^{fuz} \|\mathbf{u}^{t} - \mathbf{c}_q \|_2^2
\end{equation}

\noindent where $M$ and $Q$ are the sets of membership probability distributions and community centroid vectors, $m_{qt} \in [0, 1]$ is the probability that the $t$-th utterance belongs to the $q$-th community (with $\sum_{q=1}^{|Q|} m_{qt}=1$), $fuz$ is a parameter that controls the amount of fuzziness, $\|.\|_2$ denotes the Euclidean distance in the triplet case (we replace it with Manhattan distance $\|.\|_1$ in the siamese case), $\mathbf{u}^{t}$ is the $t$-th utterance vector, and $\mathbf{c}_q$ is the $q$-th community centroid vector.

$M$ and $Q$ are iteratively updated with equations:
\begin{align}
    m_{qt} &= \Big(\sum_{j=1}^{|Q|}\big(\frac{\|\mathbf{u}^{t} - \mathbf{c}_q \|_2}{\|\mathbf{u}^{t} - \mathbf{c}_j \|_2}\big)^{\frac{2}{fuz-1}}\Big)^{-1} \\
    \mathbf{c}_q &= \frac{\sum_{t=1}^{T} (m_{qt})^{fuz}\mathbf{u}^{t}}{\sum_{t=1}^{T} (m_{qt})^{fuz}}
\end{align}
When $fuz \to +\infty$, $\forall q\in|Q|$, $\forall t\in T$, $ m_{qt}$ tends to be equal to $\nicefrac{1}{|Q|}$, thus utterances have identical membership to each community. While when $fuz\to1$, FCM becomes equivalent to traditional $k$-means, in which $m_{qt}$ is either 0 or 1 for a given utterance $\mathbf{u}^{t}$ and community centroid $\mathbf{c}_q$. Usually in practice, $fuz=2$ \citep{schwammle2010simple}. Learning stops until the maximum number of iterations is reached or $J(M,Q)$ decreases by less than a predefined threshold. Moreover, due to its stochastic nature, we run the algorithm 20 times with different random initializations and select the run yielding the smallest objective function value. 

\end{document}